\documentclass{article}

\usepackage{microtype}
\usepackage{graphicx}
\usepackage{subfigure}
\usepackage{booktabs} % for professional tables

\usepackage[accepted]{icml2023}

\usepackage{hyperref}

\usepackage{amsmath}
\usepackage{amssymb}
\usepackage{mathtools}
\usepackage{amsthm}

\usepackage[capitalize,noabbrev]{cleveref}

\theoremstyle{plain}

\theoremstyle{definition}

\theoremstyle{remark}

\icmltitlerunning{PINA: Leveraging Side Information in eXtreme Multi-label Classification}

\usepackage{algorithm,algorithmic}
\usepackage[normalem]{ulem}
\usepackage{wrapfig}

\usepackage[symbol]{footmisc}

\begin{document}

\twocolumn[
\icmltitle{PINA: Leveraging Side Information in eXtreme Multi-label Classification via Predicted Instance Neighborhood Aggregation}

% It is OKAY to include author information, even for blind
% submissions: the style file will automatically remove it for you
% unless you've provided the [accepted] option to the icml2023
% package.

% List of affiliations: The first argument should be a (short)
% identifier you will use later to specify author affiliations
% Academic affiliations should list Department, University, City, Region, Country
% Industry affiliations should list Company, City, Region, Country

% You can specify symbols, otherwise they are numbered in order.
% Ideally, you should not use this facility. Affiliations will be numbered
% in order of appearance and this is the preferred way.
\icmlsetsymbol{intern}{*}

\begin{icmlauthorlist}
\icmlauthor{Eli Chien}{uiuc,intern}
\icmlauthor{Jiong Zhang}{amz}
\icmlauthor{Cho-Jui Hsieh}{ucla}
\icmlauthor{Jyun-Yu Jiang}{amz}
\icmlauthor{Wei-Cheng Chang}{amz}
\icmlauthor{Olgica Milenkovic}{uiuc}
\icmlauthor{Hsiang-Fu Yu}{amz}
\end{icmlauthorlist}

% \icmlaffiliation{*}{}
\icmlaffiliation{uiuc}{Department of ECE, University of Illinois Urbana-Champaign, USA}
\icmlaffiliation{amz}{Amazon, USA}
\icmlaffiliation{ucla}{Department of CS, University of California, Los Angeles, USA}

\icmlcorrespondingauthor{Eli Chien}{ichien3@illinois.edu}
% \icmlcorrespondingauthor{Firstname2 Lastname2}{first2.last2@www.uk}

% You may provide any keywords that you
% find helpful for describing your paper; these are used to populate
% the "keywords" metadata in the PDF but will not be shown in the document
\icmlkeywords{Extreme multi-label classification, Graph learning, Side information}

\vskip 0.3in
]

% this must go after the closing bracket ] following \twocolumn[ ...

% This command actually creates the footnote in the first column
% listing the affiliations and the copyright notice.
% The command takes one argument, which is text to display at the start of the footnote.
% The \icmlEqualContribution command is standard text for equal contribution.
% Remove it (just {}) if you do not need this facility.

\printAffiliationsAndNotice{$^*$This work was done during Eli Chien’s internship at Amazon, USA.}  % leave blank if no need to mention equal contribution
% \printAffiliationsAndNotice{\icmlEqualContribution} % otherwise use the standard text.

\begin{abstract}

The eXtreme Multi-label Classification~(XMC) problem seeks to find relevant labels from an exceptionally large label space. Most of the existing XMC learners focus on the extraction of semantic features from input query text. However, conventional XMC studies usually neglect the side information of instances and labels, which can be of use in many real-world applications such as recommendation systems and e-commerce product search. We propose Predicted Instance Neighborhood Aggregation (PINA), a data enhancement method for the general XMC problem that leverages beneficial side information. Unlike most existing XMC frameworks that treat labels and input instances as featureless indicators and independent entries, PINA extracts information from the label metadata and the correlations among training instances. Extensive experimental results demonstrate the consistent gain of PINA on various XMC tasks compared to the state-of-the-art methods: PINA offers a gain in accuracy compared to standard XR-Transformers on five public benchmark datasets. Moreover, PINA achieves a $\sim 5\%$ gain in accuracy on the largest dataset LF-AmazonTitles-1.3M. Our implementation is publicly available \url{https://github.com/amzn/pecos/tree/mainline/examples/pina}.
\end{abstract}

\section{Introduction}

Many real-world applications, such as e-commerce dynamic search advertising~\cite{prabhu2014fastxml,prabhu2018parabel}, semantic matching~\cite{chang2021extreme}, and open-domain question answering~\cite{Chang2020Pre-training,lee2019latent}, can be formulated as eXtreme Multi-label Classification (XMC) problems. Given a text input, XMC aims to predict relevant labels from a label collection of extremely large size $L$. The scale of $L$, which is often of the order of millions, makes designing accurate and efficient XMC models arduous.

\begin{figure}[!t]
    \centering
    \includegraphics[width=\linewidth]{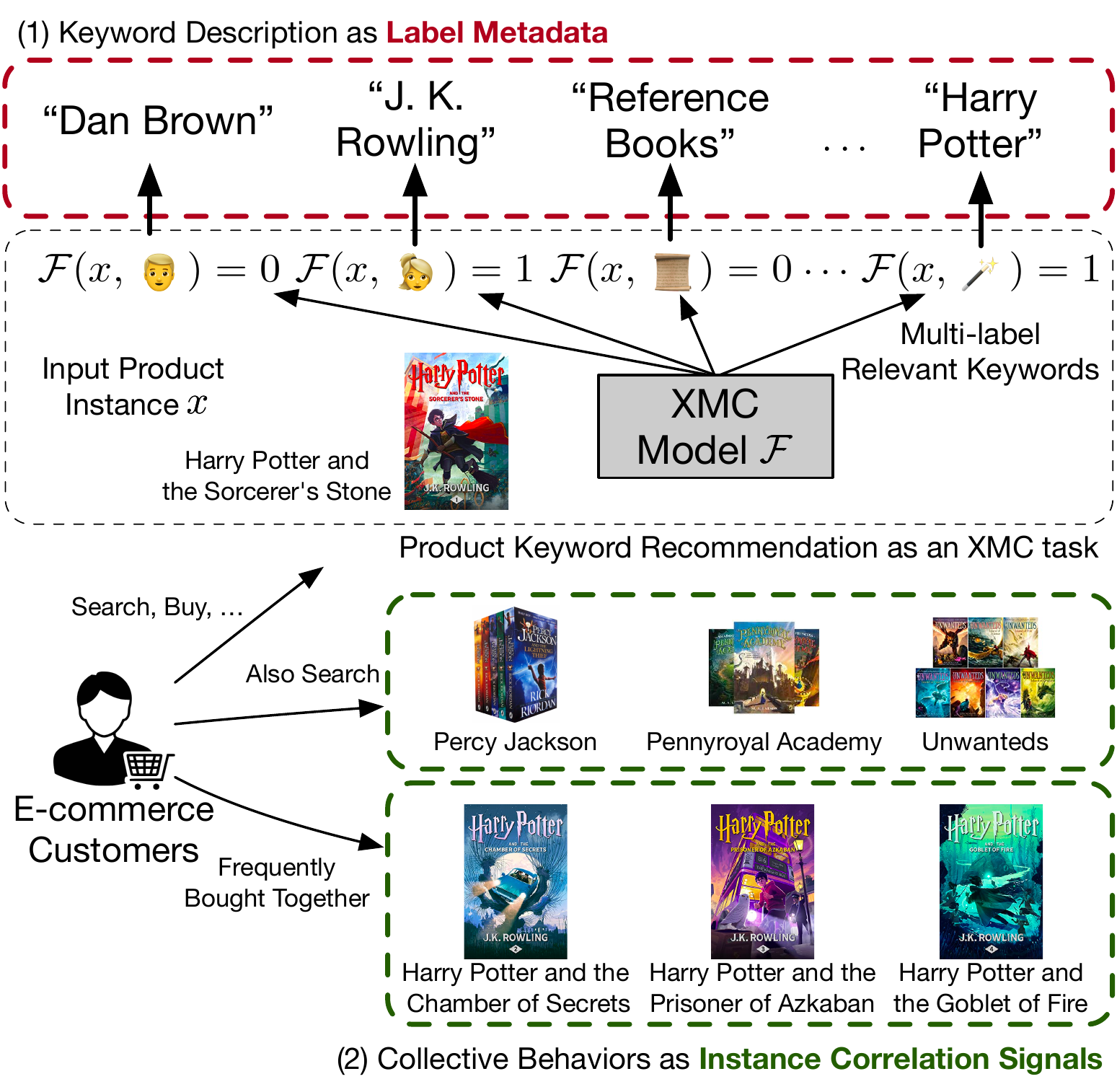}
    \caption{Illustration of two types of side information, including (1) label metadata and (2) instance correlation signals, based on an example XMC task that recommends relevant keywords (labels) for input products (instances) in E-commerce. Specifically, the text descriptions of keywords serve as label metadata while customer behaviors collectively provide instance correlation signals.}
    \label{fig:side-info}
    \vspace{-0.2cm}
\end{figure}

Despite the progress in tackling the XMC problem, most XMC solvers still only take instance features as inputs for prediction. Even though side information, such as label metadata (i.e., label text) and instance correlation signals, may be highly beneficial for the learning task, it cannot be leveraged directly. Taking product keyword recommendation as an example, 
Figure~\ref{fig:side-info} illustrates two types of side information. For label metadata, the standard XMC formulation treats labels as identifiers and ignores their text descriptions~\cite{you2019attentionxml,babbar2019data}. More precisely, while recent XMC solutions such as XR-Linear~\cite{yu2020pecos} and XR-Transformer~\cite{zhang2021fast} have exploited the correlations among labels to generate label partitions or hierarchical label trees, they do not use label text features. Instead, they construct label embeddings via aggregation of positive instance features. Recent works~\cite{mittal2021decaf,dahiya2021siamesexml} have also demonstrated that using label text features is beneficial for the XMC problem, leading to state-of-the-art results on datasets containing label text information. Moreover, instance correlation signals based on the collective behaviors of customers are also ignored in the standard XMC formulation. For example, the co-purchase signal from Amazon is now used as a benchmark graph dataset for node classification problems~\cite{chiang2019cluster,hu2020open}.
Beyond e-commerce, the idea of leveraging side information is universal and can be applied to XMC tasks in diverse fields, such as disease descriptions and cross-disease statistics in medical diagnosis~\cite{almagro2020icd}. Hence, it is of critical importance and expected to be widely impactful to enable side information inclusion into XMC models and thereby enhance prediction quality.

In the recent graph learning literature, \citealt{chien2021node} have bridged the gap between XMC and neighborhood prediction. Intuitively, the XMC label matrix can be described as a biadjacency matrix of a bipartite graph connecting instances and labels. As shown in Figure~\ref{fig:XMC_graph}, the XMC task leads to the problem of  predicting the neighborhood of each instance, which is termed the neighborhood prediction task~\cite{chien2021node}. This work clearly illustrates the point that graph learning techniques can be useful in addressing XMC tasks. One standard operation to enhance the performance of graph learning methods is graph convolution~\citep{kipf2017semisupervised}, or message passing~\citep{gilmer2017neural}. The idea is to aggregate the neighborhood features, which implicitly encode the graph topological information. The graph convolution operation has by now been successfully used in various graph learning methods, including generalized PageRank~\citep{li2019optimizing}, Graph Neural Networks (GNNs)~\citep{hamilton2017inductive,velickovic2018graph,chien2020adaptive,chien2021node} and hypergraph learning~\citep{chien2019hs,chien2021landing,chien2021you}. This work asserts that aggregating neighborhood features can also be beneficial for XMC.

Motivated by the connection between XMC and neighborhood prediction, we propose Predicted Instance Neighborhood Aggregation, PINA, to allow XMC methods such as XR-Transformers to leverage the aforementioned side information in a data enhancement manner. Our contributions can be summarized as follows:
\begin{enumerate}
    \item We introduce PINA, a data enhancement method that allows XMC models to leverage two types of side information, label metadata and instance correlation signal in a unified manner.
    \item On five public benchmark datasets where the side information is label metadata, we compare PINA with the state-of-the-art XMC model, XR-Transformer. PINA consistently beats classical XR-Transformers and achieves roughly a $5\%$ gain in accuracy on the largest dataset LF-AmazonTitles-1.3M. Moreover, XR-Transformer enhanced by the PINA technique is shown to outperform all previous published results. 
    \item We test PINA on the industrial scale proprietary dataset containing millions of instances and labels, where the side information is of the form of instance correlation signals. PINA provides a $3.5\%$ relative improvement in accuracy compared to the baseline XMC method.
\end{enumerate}
In summary, our approach consistently improves XR-Transformer on public benchmark datasets~\cite{Bhatia16} when the side information is label text. We achieve new state-of-the-art results on the public benchmark datasets with a significant gain, and also observe performance gains brought forth by PINA on proprietary datasets when the side information is in the form of instance correlation signals.

\begin{figure*}[t]
    \centering
    \includegraphics[trim={0cm 8cm 5.5cm 0},clip,width=0.8\linewidth]{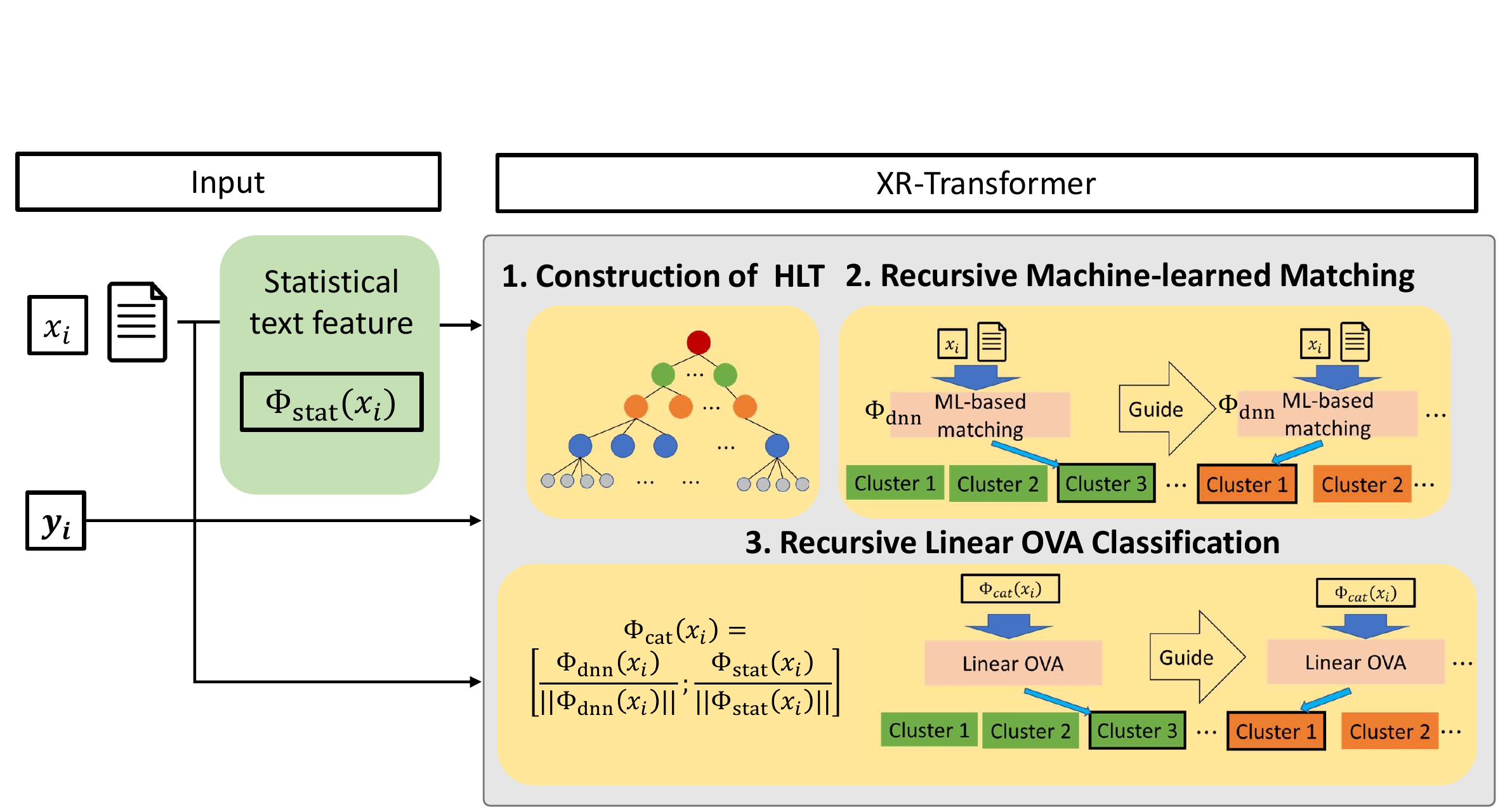}
    \vspace{-0.2cm}
    \caption{Illustration of the simplified XR-Transformer model. First, the model uses statistical text features (i.e. BoW or TF-IDF) and training labels to build the hierarchical label tree (HLT). Note that each layer of the HLT itself represents an XMC problem. Second, it trains a transformer $\Phi_{\text{dnn}}$ from the root to the leaves in a recursive manner. Third, it concatenates the statistical text feature $\Phi_{\text{stat}}(x)$ and transformer feature $\Phi_{\text{dnn}}(x)$ for learning linear one-versus-all (OVA) classifiers recursively.}
    \label{fig:XRT}
\end{figure*}

% Related work here
\vspace{-0.2cm}
\section{Related Work}
\subsection{Extreme multi-label classification}
Pioneering works on XMC adopt static input text representations and focus on the handling of
extremely large label space. Treating labels as being binary, OVA architectures such as DiSMEC~\cite{babbar2017dismec} and PPDSparse~\cite{yen2017ppdsparse} require carefully designed parallel training algorithms to handle an enormously large number of labels. Even though these methods encourage model sparsity through weight truncation, the linear inference time with respect to the output space would still make them impractical to handle millions of labels. To address this issue, some works have focused on shortlisting candidate labels to achieve sub-linear training and inference complexity. One line of study focuses on partitioning label spaces. Tree-based methods~\cite{choromanska2015logarithmic, daume2017logarithmic} divide the label space recursively into hierarchical label trees and therefore come with logarithmic inference times. More recent works such as Parabel~\cite{prabhu2018parabel}, Xtext~\cite{wydmuch2018no}, Bonsai~\cite{khandagale2020bonsai}, NapkinXC~\cite{jasinska2020probabilistic,jasinska2021online} and XR-Linear~\citep{yu2020pecos} use the tree-partitioning architecture. Approximate nearest neighbor search (ANNS) is another method that adopts shortlisting the candidate labels in XMC. Instead of restricting the search space by partitioning, methods like AnnexML~\cite{tagami2017annexml}, SLICE~\cite{jain2019slice} and GLaS~\cite{guo2019breaking} accelerate the search in the original label space with pre-build label indexing~\citep{malkov2020hnsw} or product quantization~\citep{guo2020accelerating}. 

% \Eli{To Jiong: Can you address the comments about GLaS (on regularization) and fix the issue of 2019 works improving 2020 works?}

\subsection{Deep learning based methods}
Recent works on deep learning based XMC models adopt different neural network architectures to extract semantic features and have demonstrated better performance than methods using statistical features only, such as bag-of-words (BoW) and Term Frequency-Inverse Document Frequency (TF-IDF). Methods that use shallow networks such as XML-CNN~\cite{liu2017deep} and AttentionXML~\cite{you2019attentionxml} employ CNN and BiLSTM to directly extract semantic representation from input text. On the other hand, token embedding based methods~\cite{medini2019extreme,dahiya2021deepxml,mittal2021decaf,saini2021galaxc,mittal2021eclare} use shallow networks to combine pre-trained token embeddings into input sequence representations. Despite having limited capacity to capture semantic meanings, token embedding based methods still offer good performance on short-text applications (search queries, product titles, and document keywords).

With the development of Transformer models~\cite{devlin2018bert,liu2019roberta,yang2019xlnet}, new state of the art results have been established on XMC benchmarks through fine-tuning the Transformer encoders on the downstream XMC tasks~\cite{chang2020xmctransformer,ye2020pretrained,jiang2021lightxml}. X-Transformer \cite{chang2020xmctransformer} and LightXML~\cite{jiang2021lightxml} fine-tune the transformer encoders on a simplified XMC problem, where each new label is induced by a cluster of the original labels. XR-Transformer~\cite{zhang2021fast} adopts the tree-based label partitioning approach and fine-tunes the transformer encoder on multi-resolution objectives.

\subsection{XMC with label features}
While most traditional XMC architectures treat labels as featureless identifiers, a recent study shows that taking label textural descriptions into consideration enhances the performance of XMC models~\cite{dahiya2021deepxml}. Following this line of thought, methods such as  GalaXC~\cite{saini2021galaxc}, ECLARE~\cite{mittal2021eclare} and SiameseXML~\cite{dahiya2021siamesexml} were put forward. While these methods obtained reasonable performance improvement by using label text especially when input texts are short, most of them make the assumption that the instance label bipartite graph is homophilic by using bi-encoders for candidate set retrieval (i.e. similar nodes are likely to have edges). While this is true for most XMC benchmark datasets, it does not hold in many real-world applications. For instance, complementary product recommendations in e-commerce would prefer to recommend accessories to a user who just bought a smartphone rather than yet another smartphone. Also, none of these works consider the instance correlation signals as our work.
% Related work here

\begin{figure*}[t]
    \centering
    \includegraphics[trim={1cm 6.5cm 1cm 0},clip, width=0.8\linewidth]{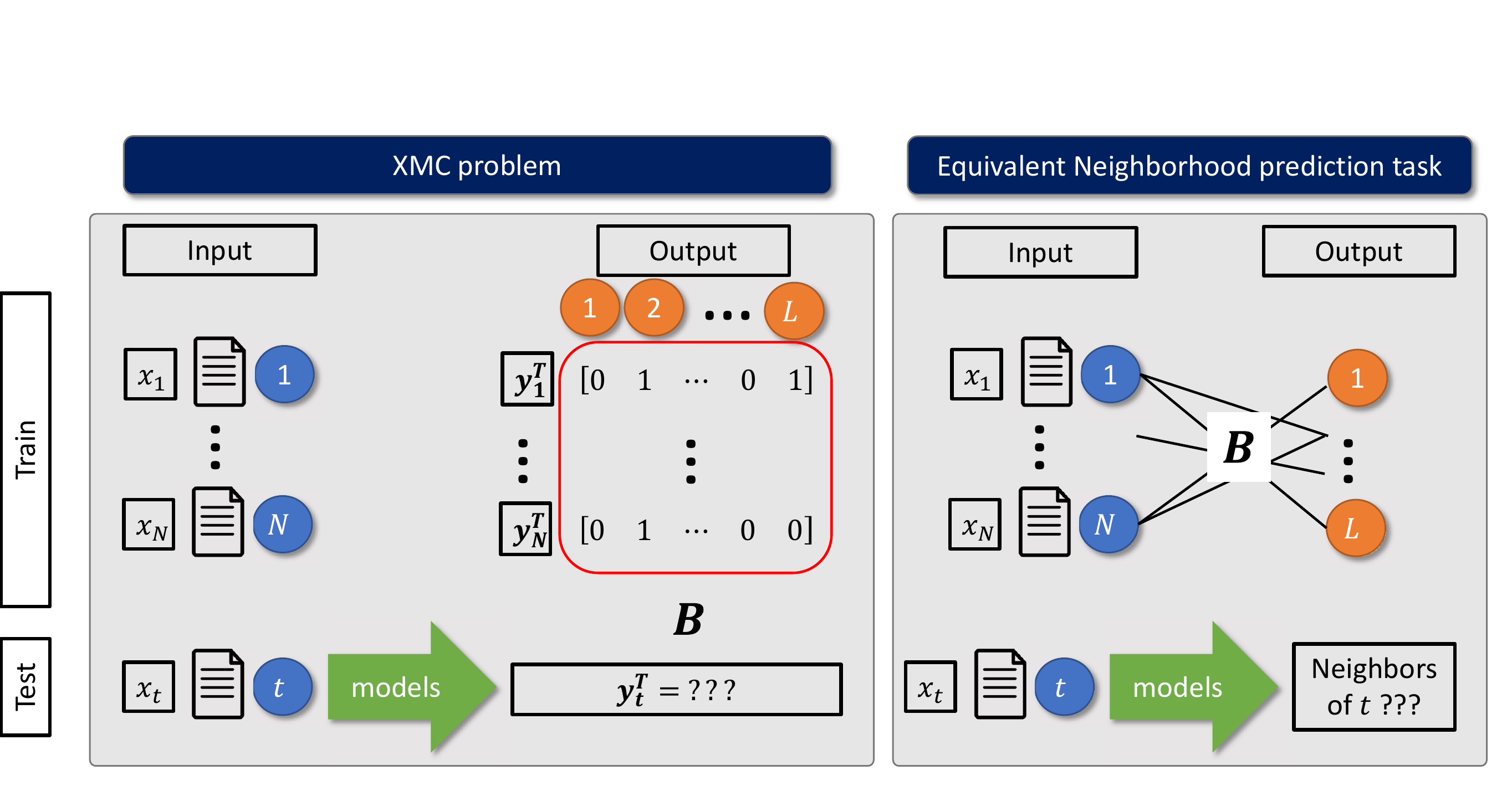}
    \vspace{-0.2cm}
    \caption{Equivalence of the XMC problem and neighborhood prediction problem. Blue nodes correspond to instances and orange nodes correspond to labels. Note that the multi-label vectors $\{\mathbf{y}_i\}$ can be viewed as the rows of biadjacency matrix $\mathbf{B}$, which characterize the edges in the graphs on the right. Hence, predicting the multi-label $\mathbf{y}_i$ is equivalent to predicting the neighborhood of blue node $i$ in the graph on the right.}
    \label{fig:XMC_graph}
\end{figure*}

\section{Preliminaries}

\begin{figure*}[t]
    \centering
    \includegraphics[trim={0cm 2.2cm 0cm 2.5cm},clip,width=0.8\linewidth]{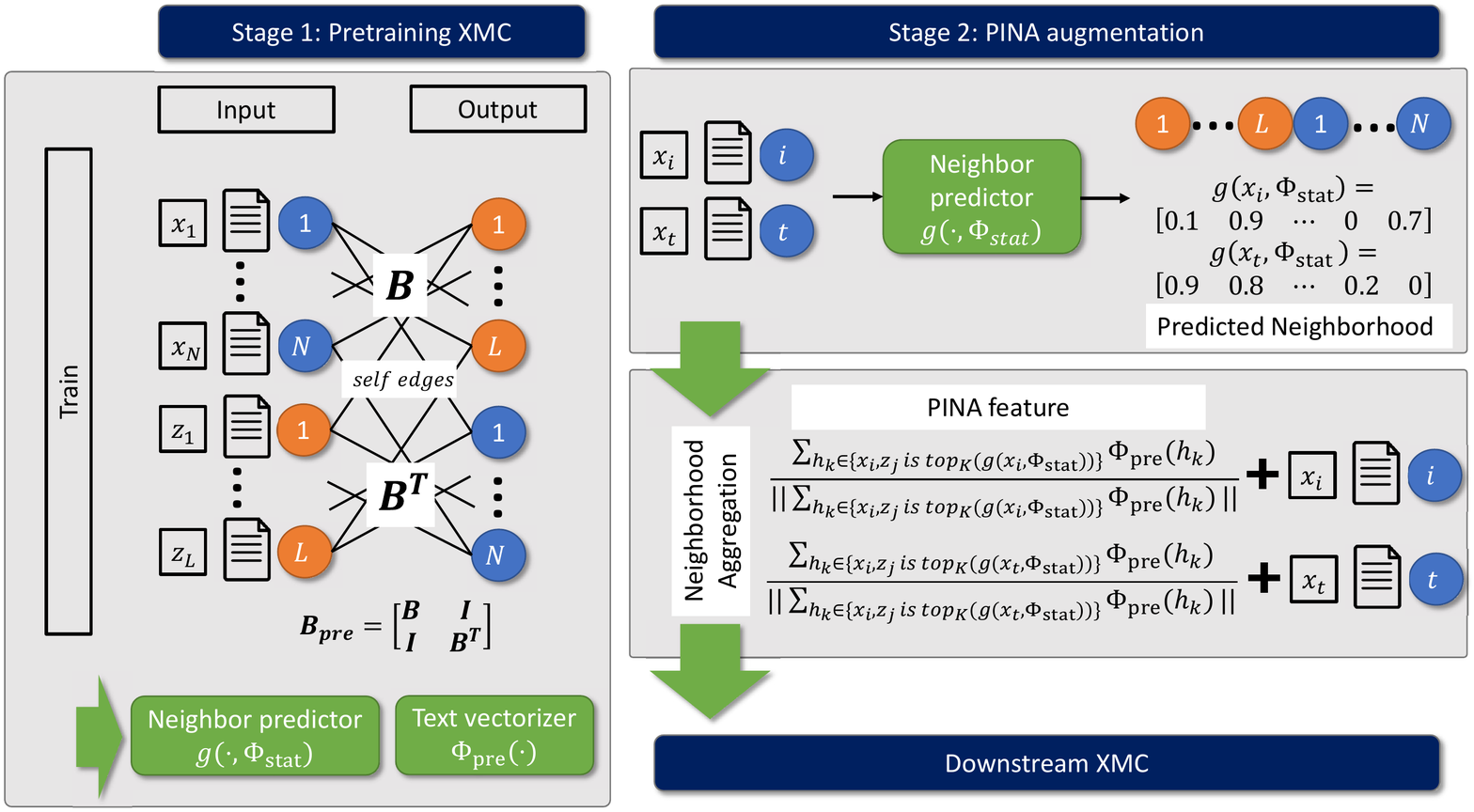}
    \vspace{-0.2cm}
    \caption{Illustration of the two-stage PINA method. At stage $1)$, we construct a pretraining biadjacency matrix $\mathbf{B}_{pre}$ using only the \emph{training data}. Since we still have an XMC problem, we can train an XMC learner as the neighbor predictor $g(\cdot,\Phi_{\text{stat}})$ and obtain its corresponding text vectorizer $\Phi_{\text{pre}}(\cdot)$ as well. At stage $2)$, we first use the pretrained neighbor predictor $g(\cdot,\Phi_{\text{stat}})$ to extract the most relevant (top $K$) nodes among the output space of the pretraining task. Then we apply the pretrained text vectorizer $\Phi_{\text{pre}}(\cdot)$ to obtain the numerical features for both instances and labels. Finally, we perform normalized neighborhood aggregation to obtain the PINA augmented features, which can then be used in downstream XMC.}
    \label{fig:PINA}
\end{figure*}

Assume that we are given a training set $\{x_i,\mathbf{y}_i\}_{i=1}^N$ where $x_i \in \mathcal{D}$ is the $i^{th}$ input instance text feature and $\mathbf{y}_i \in \{0,1\}^L$ is the one hot label vector with $y_{i,l} = 1$ indicating that label $l$ is relevant to instance $i$. The standard goal of XMC is to learn a function $f : \mathcal{D}\times [L] \mapsto \mathbb{R}$, such that $f(x,l)$ indicates the ``mutual relevance'' between $x$ and $l$. The standard way to compute this relevance is to use the one-versus-all (OVA) strategy:
\begin{align}\label{eq:XMC_OVA}
    f(x,l) = \mathbf{w}_l^T \Phi(x);\;l\in[L],
\end{align}
where $\mathbf{W} = [\mathbf{w}_1,\ldots,\mathbf{w}_L]\in \mathbb{R}^{d\times L}$ are learnable weight vectors and $\Phi: \mathcal{D}\mapsto \mathbb{R}^d$ is the text vectorizer. The function $\Phi(\cdot)$ can be obtained by either statistical methods such as BoW and TF-IDF models, or through the use of deep learning models with learnable weights. In practice, directly training with OVA is prohibitive when $L$ is large. This is due to not only the underlying $O(L)$ time complexity, but also due to severe label sparsity issues inherent to long-tailed label distributions~\cite{chang2020taming,zhang2021fast}.

\textbf{XR-Transformers. }We start by briefly introducing the state-of-the-art XMC method: XR-Transformers~\cite{zhang2021fast}. A simplified illustration of it is given in Figure~\ref{fig:XRT}. The first step is to leverage the statistical text vectorizer $\Phi_{\text{stat}}(x)$ and training labels $\mathbf{y}$ to construct the label representation $\mathbf{Z}\in \mathbb{R}^{L\times d}$ (which should not be confused with the label text feature $\{z_l\}_{l=1}^L$). XR-Transformers adopt the Predicted Instance Feature Aggregation (PIFA) strategy for label representations, which is further used to construct the hierarchical label tree (HLT) via hierarchical $k$-means clustering,
% \vspace{-0.05cm}
\begin{align}\label{eq:PIFA}
    \text{(PIFA)}\;\quad & \mathbf{Z}_l = \frac{\sum_{i:y_{il}=1}\Phi_{\text{stat}}(x_i)}{\|\sum_{i:y_{il}=1}\Phi_{\text{stat}}(x_i)\|}\;\forall l \in [L].
\end{align}

Note that each level of the HLT gives rise to an XMC problem. The second step is to train the Transformer models $\Phi_{\text{dnn}}$, such as BERT~\cite{devlin2018bert}, recursively from root to leaves. In the third step, the XR-Transformer concatenates both the statistical feature $\Phi_{\text{stat}}(x)$ and Transformer embedding $\Phi_{\text{dnn}}(x)$ to arrive at the final feature $\Phi_{\text{cat}}(x)$. It also trains linear OVA classifiers~\eqref{eq:XMC_OVA} based on HLT recursively to generate the final prediction. Through the use of HLT, one can not only reduce the time complexity from $O(L)$ to $O(\log(L))$, but also alleviate the label sparsity issue~\cite{zhang2021fast}.

\textbf{XMC with label text. }Consider the scenario where the label text $\{z_l\}_{l=1}^L$ is available as side-information, where $z_l\in \mathcal{D}$ is the label text of label $l$. One can observe that standard XMC approaches, such as XR-Transformers, cannot leverage this information directly. While it is possible to use the label text to improve the construction of the HLT, the learnable text vectorizer $\Phi_{\text{dnn}}$ itself cannot leverage the label text information. PINA, as we show, enables XMC learners to leverage label text information in a data enhancement manner, where the learnable text vectorizer $\Phi_{\text{dnn}}$ can also perform training with the label texts.

\textbf{XMC with instance correlation signal. }In keyword recommendation problems, researchers aim to predict the most relevant keywords for each product. Keyword recommendation is an example of an XMC problem. In this scenario, instances (products) correlation signals are also available from the customer behavioral data, such as those pertaining to the ``frequently bought together'' category. This type of side information provides us with beneficial information about the instances. Unfortunately, it is not clear how to leverage the instances correlation signals within the standard XMC problem solvers. PINA makes use of this side information in a data enhancement way similar to what is done with the label text.

\subsection{The XMC problem and the neighborhood prediction problem}

To understand the key idea behind our approach, we have to describe the relationship between the XMC problem and the neighborhood prediction problem first described in the graph learning literature. Recently,~\citealt{chien2021node} revealed the equivalence of the XMC problem and the neighborhood prediction problem in graph learning. Let $G=(V_{\text{in}},V_{\text{out}},E)$ be a directed bipartite graph, where $V_{\text{in}}=[N]$ and $V_{\text{out}}=[L]$ are the input and output node sets, respectively, while $E \subseteq V_{\text{in}}\times V_{\text{out}}$ is the edge set. A common way to characterize the edge relations is to use a biadjacency matrix $\mathbf{B}\in \{0,1\}^{N\times L}$, where $B_{ij} = 1$ if and only if $(i,j)\in E$. The goal of the neighborhood prediction problem is to predict the $i^{th}$ row of $\mathbf{B}$ via the node attributes of node $i$. Since the $i^{th}$ row of $\mathbf{B}$ is just a vector in $\{0,1\}^{1\times L}$ (i.e., a binary vector), it can also be viewed as a multi-label $\mathbf{y}_i$. See Figure~\ref{fig:XMC_graph} for a pictorial illustration. 

One standard operation in graph learning is graph convolution~\cite{kipf2017semisupervised}, where the key idea is to gather the attributes of the neighborhood of a node to enhance its ego node features. It has been proven to be  effective for many graph tasks, including node classification~\cite{kipf2017semisupervised,hamilton2017inductive,velickovic2018graph,chien2020adaptive}, link prediction~\cite{zhang2021labeling,zhang2018link} and graph classification~\cite{xu2018how,zhang2021nested}. Our proposed method -- PINA -- is motivated by the connection to the neighborhood prediction task and the graph convolution operation, which we describe in the next section.

\section{PINA: Predicted Instance Neighborhood Aggregation}\label{sec:PINA-main}

For simplicity, we mostly focus on the setting where side information is of the form of label text. The case of side information being of the form of instance correlation signals can be treated similarly. A detailed discussion regarding how to apply PINA with instance correlation signals is available in Section~\ref{sec:other_side_info}.

We propose PINA to allow XMC learners such as XR-Transformers to make use of label text information in a \emph{data enhancement} manner. Due to the equivalence of XMC and neighborhood prediction, a naive way of including label attributes is via neighborhood aggregation. However, there are several issues preventing us from applying this idea directly. First, one can only apply the average operation on numerical features instead of raw text. Ideally, we have to fine-tune the text vectorizer $\Phi_{\text{dnn}}$ with both instance and label text features during the training phase. However, the XMC formalism (Figure~\ref{fig:XMC_graph}) does not treat label text as an input, which is suboptimal. Second, the neighborhood relation is defined using labels $\mathbf{y}_i$, which are \emph{unknown for test instances}. See Figure~\ref{fig:XMC_graph} for an illustration. Thus, we cannot apply neighborhood aggregation directly even though we are equipped with the bipartite graph underlying the XMC problem. We describe next the high-level ideas how to resolve these issues.

\textbf{Lack of knowledge about neighborhoods for test instances. }In order to resolve this issue, we propose to pretrain a neighborhood predictor $g$. Instead of using the exact neighbors (i.e. ground truth multi-labels), we generate predicted neighbors via $g$. This allows us to generate neighbors for both \emph{train and test} instances. Note that pretraining $g$ only leverages training data (which includes both labels and instances).

\textbf{The transformer text vectorizer $\Phi_{\text{dnn}}$ does not involve label text. }In order to resolve this issue, we propose a pretraining XMC task that also takes label text as input. More specifically, the input and output space of our pretraining task contains both instances and labels. See the illustration of the proposed pretrained XMC in Figure~\ref{fig:PINA}. Hence, our pretrained text vectorizer $\Phi_{\text{pre}}$ is trained with both instance text and label text. This resolves the issue of not being able to include the label text in standard XMCs.

\subsection{A detailed description of PINA}
We implemented PINA as a two-stage method, described in Figure~\ref{fig:PINA}. The pseudo-code of the PINA augmentation and pretraining process are listed in the Appendix~\ref{apx:pseudo-code}. The first stage is the pretraining phase, where we design a pretraining task to learn a neighbor predictor $g(\cdot,\Phi_{\text{stat}})$ via a base XMC learner (e.g., XR-transformer). Note that the pretraining task is also an XMC problem, but both instances and label text are treated as inputs and both the input and output space contain instance and label nodes. The edges are defined by multi-label relations $\{\mathbf{y_i}\}_{i=1}^N$ in an undirected manner. We also add edges from all instances and label nodes in the input space to their output space counterpart. More specifically, we construct $\mathbf{B}_{pre}$ as described in Figure~\ref{fig:PINA}. Recall that $\mathbf{B}\in \{0,1\}^{N \times L}$ is obtained by training the multi-labels $\{\mathbf{y}_i\}_{i=1}^N$ and $\mathbf{I}$ represents an identity matrix of appropriate dimensions. Hence, in our pretraining XMC problem, we aim to predict the $i^{th}$ row of $\mathbf{B}_{pre}$ using $x_i$ when $i\in [N]$ and $z_{i-N}$ when $i = N+1,N+2,\ldots,N+L$. This allows both the label and instance text to be observed during the pretraining phase. We consequently obtain the corresponding text vectorizer $\Phi_{\text{pre}}$ to generate numerical features for both the labels and instances. 

The second stage is the PINA augmentation phase, during which we leverage the pretrained neighborhood predictor $g(\cdot,\Phi_{\text{stat}})$ and text vectorizer $\Phi_{\text{pre}}(\cdot)$ to augment the instance features. We first predict the most relevant nodes among the output space of the pretraining stage as neighbors for both training and \emph{test} instances via our pretrained neighbor predictor $g(\cdot,\Phi_{\text{stat}})$. More specifically, we obtain the neighborhood prediction vector $g(x_i,\Phi_{\text{stat}})\in [0,1]^{1\times L}$ and zero out all but the top $K$ largest values $\mathbf{P}_i = top_K(g(x_i,\Phi_{\text{stat}}))$. Then we perform neighborhood aggregation on the numerical features obtained from $\Phi_{\text{pre}}$ accordingly, which results in PINA features. See lines $3-8$ in Algorithm~\ref{alg:PINA_augmentation} for PINA feature extraction of each instance. The augmented features are fed to the next XMC learner for solving the downstream XMC task. 

\subsection{Applying PINA to instance correlation signals} \label{sec:other_side_info}
We describe next how to apply PINA to instance correlation signals. In this case, the instance correlation can be formulated as an instance-to-instance graph (i.e., I2I graph). Similarly to the construction rules of the Amazon co-purchase graph benchmarking dataset known from the graph learning literature, a link $(i,j)$ exists if and only if the instance correlation signal between $i$ and $j$ is larger than a threshold. We thus capture the instance correlation signal by the (bi)adjacency matrix $\mathbf{B}^\prime$.  

We then use $\mathbf{B}^\prime$ directly to formulate our pretraining XMC. More specifically, one can choose $\mathbf{B}_{pre} = \mathbf{B}^\prime$ in Stage 1 of Figure~\ref{fig:PINA} to obtain the neighbor predictor and text vectorizer. Note that the set of instances considered in the downstream application need not be identical to instances in the I2I graph. We only require their text domains to be the same (i.e., all texts associated with instances are product descriptions). This is due to the fact that we can still obtain the predicted neighbors among instances in the I2I graph for each instance in the downstream application (and similar to the reason why PINA applies to test data). The remainder of the PINA pipeline is as described in Figure~\ref{fig:PINA}.

\begin{table}[t]
\centering
\caption{Public benchmark dataset statistics: $N_{train}, N_{test}$ refer to the
    number of instances in the training and test sets, respectively;
    $L$: the number of labels. $\bar{L}$: the average number of positive labels per
    instance; $\bar{n}$: average number of instances per label; $d_{BoW}$:
    the Bag-of-Word feature dimension.
}
\label{tab:data_stats}
\scriptsize
\setlength{\tabcolsep}{2pt}
\vspace{0.1 in}
\begin{tabular}{@{}ccccccc@{}}
\toprule
 & $d_{BoW}$ & $L$ & $N_{train}$ & $N_{test}$ & $\bar{n}$ & $\bar{L}$ \\
 \midrule
LF-Amazon-131K      & 80,000 & 131,073 & 294,805   & 134,835 & 5.15  & 2.29 \\
LF-WikiSeeAlso-320K & 80,000 & 312,330 & 693,082   & 177,515 & 4.67  & 2.11 \\
LF-Wikipedia-500K   & 500,000 & 501,070 & 1,813,391 & 783,743 & 24.75 & 4.77 \\ 
LF-Amazon-1.3M   & 128,000 & 1,305,265 & 2,248,619 & 970,273 & 28.24 & 22.20 \\\bottomrule
\end{tabular}
\end{table}

\section{Experimental Results}
% ========== Table for LF results ===========
\begin{table*}[t]
  \centering
    \caption{Main result on label text XMC benchmark datasets. Bold font refers to the best result. Superscripts $^\dagger$ and $^\star$ indicate the results are taken from DECAF paper~\cite{mittal2021decaf} and SiameseXML~\cite{dahiya2021siamesexml} respectively.}
    \label{tab:LF-results}
    \vspace{0.1 in}
    \resizebox{1.0\textwidth}{!}{
    \begin{tabular}{c|cccc|cccc}
\hline
Methods &
  \textbf{P@1} &
  \textbf{P@3} &
  \multicolumn{1}{c|}{\textbf{P@5}} &
  \textbf{Train Time (hrs)} &
  \textbf{P@1} &
  \textbf{P@3} &
  \multicolumn{1}{c|}{\textbf{P@5}} &
  \textbf{Train Time (hrs)} \\ \hline
                        & \multicolumn{4}{c|}{LF-Amazon-131K}                                  & \multicolumn{4}{c}{LF-WikiSeeAlso-320K}            \\ \hline
DECAF$^\dagger$         & 42.94 & 28.79          & \multicolumn{1}{c|}{21}             & 1.8   & 41.36 & 28.04 & \multicolumn{1}{c|}{21.38} & 4.84  \\
AttentionXML$^\dagger$  & 42.9  & 28.96          & \multicolumn{1}{c|}{20.97}          & 50.17 & 40.5  & 26.43 & \multicolumn{1}{c|}{21.38} & 90.37 \\
SiameseXML$^\star$      & 44.81 & 30.19              & \multicolumn{1}{c|}{21.94}          & 1.18  & 42.16 & 28.14     & \multicolumn{1}{c|}{21.35} & 2.33  \\
ECLARE$^\star$          & 43.56 & 29.65              & \multicolumn{1}{c|}{21.57}          & 2.15  & 40.58 & 26.86     & \multicolumn{1}{c|}{20.14} & 9.40  \\
$\text{XR-Transformer}$ & 45.61 & 30.85          & \multicolumn{1}{c|}{22.32}          & 7.9   & 42.57 & 28.24 & \multicolumn{1}{c|}{21.30} & 22.1  \\
$\text{XR-Transformer + PINA}$ &
  \textbf{46.76} &
  \textbf{31.88} &
  \multicolumn{1}{c|}{\textbf{23.20}} &
  9.8 &
  \textbf{44.54} &
  \textbf{30.11} &
  \multicolumn{1}{c|}{\textbf{22.92}} &
  28.3 \\ \hline
                        & \multicolumn{4}{c|}{LF-Wikipedia-500K}                               & \multicolumn{4}{c}{LF-Amazon-1.3M}                 \\ \hline
DECAF$^\dagger$         & 73.96 & 54.17          & \multicolumn{1}{c|}{42.43}          & 44.23 & -     & -     & \multicolumn{1}{c|}{-}     & -     \\
AttentionXML$^\dagger$  & 82.73 & \textbf{63.75} & \multicolumn{1}{c|}{\textbf{50.41}} & 221.6 & -     & -     & \multicolumn{1}{c|}{-}     & -     \\
SiameseXML$^\star$      & 67.26 & 44.82          & \multicolumn{1}{c|}{33.73}          & 7.31  & -     & -     & \multicolumn{1}{c|}{-}     & -     \\
ECLARE$^\star$          & 68.04 & 46.44          & \multicolumn{1}{c|}{35.74}          & 86.57 & -     & -     & \multicolumn{1}{c|}{-}     & -     \\
$\text{XR-Transformer}$ & 81.62 & 61.38          & \multicolumn{1}{c|}{47.85}          & 41.0  & 54.67 & 47.87 & \multicolumn{1}{c|}{42.93} & 28.2  \\
$\text{XR-Transformer + PINA}$ &
  \textbf{82.83} &
  63.14 &
  \multicolumn{1}{c|}{50.11} &
  85.0 &
  \textbf{58.33} &
  \textbf{51.06} &
  \multicolumn{1}{c|}{\textbf{46.04}} &
  39.1 \\ \hline
\end{tabular}
}
\end{table*}

% ========== Table for LF results ===========

We demonstrate the effectiveness of PINA on both public benchmark datasets with side information of the form of label text and proprietary datasets with side information of the form of instance correlation signals. We report the precision at $k$ (P@$k$) and recall at $k$ (R@$k$) as our evaluation metrics. Their definition can be found in Appendix~\ref{apx:Eval_detail}.

\subsection{Label text benchmark datasets}

\textbf{Datasets. }We consider the public benchmark long-text datasets for product-to-product recommendations as well as predicting related Wikipedia articles, taken from~\cite{Bhatia16}. The data statistics can be found in Table~\ref{tab:data_stats}. For a fair comparison with previous works, we adopt the provided BoW features as our statistical text feature in the experiments. For LF-Amazon-1.3M, where BOW features are only constructed using title text, we still use the provided BoW features~\cite{Bhatia16} and leverage the full text only as input to transformer encoders.

\textbf{Baseline methods. }We not only compare PINA with plain XR-Transformers~\cite{zhang2021fast}, but also with other previously published XMC methods that achieve state-of-the-art results for the label text XMC problem. These include ECLARE~\cite{mittal2021eclare}, DECAF~\cite{mittal2021decaf}, AttentionXML~\cite{you2019attentionxml} and SiameseXML~\cite{dahiya2021siamesexml}. For methods other than XR-Transformer, we directly take the reported numbers from the DECAF paper~\cite{mittal2021decaf} with superscript $^\dagger$ and SiameseXML paper~\cite{dahiya2021siamesexml} with superscript $^\star$. For PINA, we use XR-Transformers as our baseline XMC learners.

\textbf{Results. }The results are summarized in Table~\ref{tab:LF-results}. Compared to XR-Transformers, PINA consistently improves the performance with a significant gain across all datasets. This demonstrates the effectiveness of PINA as a data enhancement approach for leveraging label text side information. As an example, PINA improves XR-Transformer with a gain of $1$-$2\%$. Compared to previously published state-of-the-art methods, XR-Transformer + PINA outperforms SiameseXML by around $2\%$ and $2.5\%$ on LF-Amazon-131K and LF-WikiSeeAlso-320K, respectively. At the same time, XR-Transformer + PINA achieves roughly the same performance as AttentionXML on LF-Wikipedia-500K. AttentionXML is outperformed by XR-Transformer + PINA with a $4\%$-$4.5\%$ margin on LF-Amazon-131K and LF-WikiSeeAlso-320K. Moreover, AttentionXML exhibits at least a $2.5$ times larger training time compared to XR-Transformer + PINA. These results again demonstrate the superiority of the proposed PINA method in leveraging label text side information in the XMC problem. More experiment details such as significant tests are included in the Appendix~\ref{apx:Eval_detail}.

Note that none of the previously reported methods were tested on the (long text) LF-Amazon-1.3M dataset. Some where tested on the short text version LF-AmazonTitles-1.3M, which exclusively uses instance titles as text features. We have included the result on the original LF-AmazonTitles-1.3M dataset in Table~\ref{tab:LF-Title}, which show that PINA outperforms published state-of-the-art methods by a large margin.

\begin{table}[ht]
\caption{Study of PINA on LF-AmazonTitle-1.3M dataset. Bold font numbers indicate the best results.}
\label{tab:LF-Title}
\vspace{0.1 in}
\begin{tabular}{@{}c|ccc@{}}
\toprule
Methods                           & \textbf{P@1} & \textbf{P@3} & \textbf{P@5} \\ \midrule
                        & \multicolumn{3}{c}{LF-AmazonTitle-1.3M}  \\ \hline

DECAF$^\dagger$         & 50.67 & 44.49  & 40.35  \\
AttentionXML$^\dagger$  & 45.04 & 39.71  & 36.25 \\
SiameseXML$^\star$      & 49.02 & 42.72  & 38.52 \\
ECLARE$^\star$          & 50.14 & 44.09  & 40.00 \\

$\text{XR-Transformer}$                  & 50.98        & 44.49        & 40.05        \\
XR-Transformer + PINA & \textbf{55.76} & \textbf{48.70} & \textbf{43.88} \\ \bottomrule
\end{tabular}
\vspace{-0.1cm}
\end{table}

\begin{table}[t]
\caption{Ablation study of PINA on the LF-Amazon-131K dataset. Bold font indicates the best results. For PINA-naive, we use only $\mathbf{B}$ as our pretraining target in Stage 1 of PINA.}
\label{tab:LF-ablation}
\vspace{0.1 in}
\begin{tabular}{@{}cccc@{}}
\toprule
LF-Amazon-131K                           & \textbf{P@1} & \textbf{P@3} & \textbf{P@5} \\ \midrule
$\text{XR-Transformer}$                  & 45.61        & 30.85        & 22.32        \\
XR-Transformer + PINA-naïve & 43.89        & 30.43        & 22.67        \\
XR-Transformer + PINA & \textbf{46.76} & \textbf{31.88} & \textbf{23.20} \\ \bottomrule
\end{tabular}
\end{table}

\textbf{Ablation study. }In Section~\ref{sec:PINA-main} we introduced two major problems in implementing the neighborhood aggregation idea. For the problem of lacking neighborhoods for test instances, it is straightforward to understand why using predicted neighbors resolves the issue. In the ablation study that follows, we test whether letting the transformer text vectorizer $\Phi_{\text{dnn}}$ observe the label text influences the performance of the learner. Instead of pretraining with $\mathbf{B}_{pre}$, one may also pretrain the neighbor predictor with $\mathbf{B}$. Our results are listed in Table~\ref{tab:LF-ablation}. We can see that letting the transformer text vectorizer $\Phi_{\text{dnn}}$ see the label text is indeed crucial, which verifies our intuition described in Section~\ref{sec:PINA-main}. Moreover, we find that pretraining neighborhood predictors with $\mathbf{B}$ can often lead to worse results compared to applying XR-Transformer only. This further highlights the necessity of our design for PINA. Finally, we provide a qualitative analysis in Appendix~\ref{apx:qual-analysis} which further validate the necessity of the design of PINA.

\subsection{Proprietary datasets with instance correlation signal}

\textbf{Datasets. }We also conduct our experiment on a proprietary dataset pertaining to tasks similar to those illustrated in Figure~\ref{fig:side-info}. This proprietary dataset consists of millions of instances and labels. Our side information is of the form of instance correlation signals among roughly millions of instances. The instance correlation signal is aggregated similar to the case described in the Introduction.

\textbf{Settings. }Due to the large scale of the data, the requirements for small inference times and daily model updates, we choose XR-Linear~\cite{yu2020pecos} as our downstream XMC model. Nevertheless, since the pretraining step in PINA can be performed beforehand and requires less frequent updates (i.e. monthly), we can still use the XR-Transformer as our neighborhood predictor during PINA augmentation. The high-level idea of the XR-Linear model can be understood with the help of Figure~\ref{fig:XRT}, where XR-Linear does not include Step 2 (i.e., machine learned matching). It directly trains a linear OVA classifier recursively on the HLT with input statistical text features such as BoW or TF-IDF. Besides applying PINA with XR-Linear, we also conduct an ablation study. We test if our performance gain is merely a consequence of concatenating features from pretrained transformer text vectorizers or if neighborhood aggregation also plays an important role.

\textbf{Results. }We report the relative performance compared to the plain XR-Linear model. Our results are listed in Table~\ref{tab:A2Q}. One can observe that PINA once again consistently improves the performance of downstream XMC models under all reported metrics. Furthermore, our ablation study shows that the performance gain of PINA does not merely come from concatenating pretrained text features. Our neighborhood aggregation mechanism is indeed important. Notably, merely using pretrained text features can lead to worse performance in P$@10$ and R$@10$. 

\begin{table}[t]
\centering
\caption{Results on the proprietary dataset. The baseline corresponds to applying XR-Linear directly and we report the relative results compared to it in $\%$. Bold fonts indicate the best results.}
\label{tab:A2Q}
\vspace{0.1 in}
\begin{tabular}{@{}ccccc@{}}
\toprule
                                                                               & \textbf{P@1}   & \textbf{P@10} & \textbf{R@1}   & \textbf{R@10}           \\ \midrule
\begin{tabular}[c]{@{}c@{}}Use pretrained \\ text vectorizer only\end{tabular} & +0             & -1.51         & +1.67          & -1.14          \\
PINA                                                                           & \textbf{+3.45} & +0            & \textbf{+3.93} & \textbf{+0.23} \\ \bottomrule
\end{tabular}
\vspace{-0.1 in}
\end{table}

\section{Conclusion}

We proposed Predicted Instance Neighborhood Aggregation~(PINA), a data enhancement framework that allows traditional XMC models to leverage various forms of side information, such as label metadata and instance correlation signals. Motivated by the neighborhood prediction problem from the graph learning literature, PINA enriches the instance features via neighborhood aggregation similar to what graph convolutions and message-passing operations do in many graph learning tasks. We conducted experiments on both public benchmark datasets and a proprietary dataset. PINA offers consistent gains when compared to its backbone XMC models such as XR-Transformers and XR-Linear. The combination of PINA and a XR-Transformer also outperforms published state-of-the-art methods specialized for label text on all the benchmarking datasets.

% Acknowledgements should only appear in the accepted version.
\section*{Acknowledgements}
The authors thank the support from Amazon and the Amazon Conference Grant. Part of this work was funded by the NSF CIF Grant 1956384. Cho-Jui Hsieh is supported in part by NSF  IIS-$2008173$ and IIS-$2048280$.  

\bibliography{Ref}
\bibliographystyle{icml2023}

%%%%%%%%%%%%%%%%%%%%%%%%%%%%%%%%%%%%%%%%%%%%%%%%%%%%%%%%%%%%%%%%%%%%%%%%%%%%%%%
%%%%%%%%%%%%%%%%%%%%%%%%%%%%%%%%%%%%%%%%%%%%%%%%%%%%%%%%%%%%%%%%%%%%%%%%%%%%%%%
% APPENDIX
%%%%%%%%%%%%%%%%%%%%%%%%%%%%%%%%%%%%%%%%%%%%%%%%%%%%%%%%%%%%%%%%%%%%%%%%%%%%%%%
%%%%%%%%%%%%%%%%%%%%%%%%%%%%%%%%%%%%%%%%%%%%%%%%%%%%%%%%%%%%%%%%%%%%%%%%%%%%%%%
\newpage
\appendix
% \onecolumn
\section{Future directions}
Although we focused on combining PINA with XR-Transformers, it is worth pointing out that PINA can also be used in conjunction with other XMC methods. Recall that the neighborhood predictor $g$ in PINA can be modeled by another XMC method and that one can make use of the pretrained text vectorizer $\Phi_{\text{pre}}$ therein. It is also an interesting question to test the performance of PINA combined with other XMC methods.

\section{Potential Social Impact}
Our method allows XMC solutions such as XR-Transformer to leverage various side information, including label text and instance correlation signals. This can have positive social impacts such as improving search results for minorities, where XMC solvers usually have worse performance due to less data abundance. Since our method is a general approach to the XMC problem, we do not aware of any negative social impact of our work.

\section{Additional details on proprietary datasets}
We subsampled roughly 2 million instances and 3 million labels. The label matrix contains roughly 7 million positives. The underlying task is e-commerce keyword recommendation and the side information is similar to the co-purchase signal that we mentioned in our paper.

\section{Computational Environment}

All experiments are conducted on the AWS p3dn.24xlarge instance, consisting of 96 Intel Xeon CPUs with 768 GB of RAM and 8 Nvidia V100 GPUs with 32 GB of memory each.

\section{Hyperparameters of PINA and XR-Transformers}

For the PINA augmentation stage, we aggregate the top-$5$ neighbors (with respect to the highest prediction score) according to the output of the neighborhood predictor $g$. We directly use the normalized prediction score as edge weights for the neighborhood aggregation process. For the pretrained neighborhood predictor $g$ and downstream XR-Transformer, we adopt the default hyperparameters provided by the original XR-Transformer repository~\footnote{\url{https://github.com/amzn/pecos/tree/mainline/examples/xr-transformer-neurips21}} with some slight modifications. More specifically, we first choose the hyperparameters from datasets that are ``closest'' to our datasets. For example, we adopt the hyperparameters used for ``amazon-670k'' from~\citealt{zhang2021fast} for our ``LF-Amazon-131K'' dataset. We adopt the hyperparameters used for ``wiki-500k'' from~\citealt{zhang2021fast} for our ``LF-WikiSeeAlso-320K'' and ``LF-Wikipedia-500K'' datasets. We adopt the hyperparameters used for ``amazon-3m'' from~\citealt{zhang2021fast} for our ``LF-Amazon-1.3M'' dataset. For pretraining the neighborhood predictor $g$, we only change the HLT-related parameters in order to accommodate different sizes of the output space. All hyperparameters are provided in our code for reproducibility.

\section{Evaluation Details}\label{apx:Eval_detail}
Note that we follow previous works such as ECLARE~\cite{mittal2021eclare} and SiameseXML~\cite{dahiya2021siamesexml} to post-process the test data and predictions. We use the code from ECLARE for ``Reciprocal-pair Removal''. This procedure removes instance-label pairs that are reciprocal, meaning that the two instances are also each others labels. See the XMC repository website~\footnote{\url{http://manikvarma.org/downloads/XC/XMLRepository.html\#ba-pair}} for more details.

\textbf{Definition of precision and recall at $k$. }We report the precision at $k$ (P@$k$) and recall at $k$ (R@$k$) as our evaluation metrics. Their formal definitions are as follows:
\begin{align}
  \text{P@k}&=\frac{1}{k}\sum_{l=1}^k y_{rank(l)}\\
  \text{R@k}&=\frac{1}{\sum_{i=1}^L y_i}\sum_{l=1}^k y_{rank(l)}
\end{align}
where $y\in\{0,1\}^L$ is the ground truth label and $rank(l)$ is the index of the $l$-th highest predicted label.

\textbf{Significant Tests.} We conduct significant tests to verify the improvements by leveraging PINA. In particular, we calculated the test instance-wise $P@1,3,5$ for XR-Transformer and XR-Transformer+PINA and perform t-test on each one of these metrics. As shown in table~\ref{tab:t-test},  the gain of XR-Transformer+PINA over XR-Transformer is indeed significant with p-value less than $8.5\times 10^{-4}$ for all precision metrics in table~\ref{tab:LF-results}.

\begin{table}[h]
\caption{P-values of significant test between XR-Transformer and XR-Transformer+PINA on instance-wise metrics.}
\label{tab:t-test}
\vspace{0.1 in}
\resizebox{0.48\textwidth}{!}{
\begin{tabular}{@{}cccc@{}}
\toprule
        & \textbf{p-value:P@1} & \textbf{p-value:P@3} & \textbf{p-value:P@5} \\ \midrule
LF-Amazon-131K       & $8.5\times 10^{-4}$ & $1.8\times 10^{-8}$ & $1.8\times 10^{-14}$ \\
LF-WikiSeeAlso-320K  & $3.8\times 10^{-15}$ & $1.2\times 10^{-34}$ & $1.7\times 10^{-57}$ \\
LF-Wikipedia-500K    & $6.1\times 10^{-28}$ & $1.0\times 10^{-113}$ & $1.4\times 10^{-179}$ \\
\bottomrule
\end{tabular}}
\end{table}

\section{Qualitative analysis}\label{apx:qual-analysis}
\begin{table*}[t]
\caption{Qualitative analysis of predicted results on the LF-Amazon-131K dataset. Bold fonts indicate correct predictions while italic fonts represent the ``helpful'' neighbors predicted by the neighborhood predictor $g$ of PINA.}
\label{tab:exp_analysis}
\scriptsize
\vspace{0.1 in}
\begin{tabular}{@{}cccc@{}}
\toprule
Instance title: &
  \multicolumn{3}{c}{Himalayan Cats (Barron's Complete Pet Owner's Manuals)} \\ \midrule
Ground Truth (unorderly) &
  \multicolumn{3}{c}{Guide to Owning Himalayan Cat; Guide to Owning a Persian Cat; The Complete Book of Cat Breeding} \\ \midrule
\begin{tabular}[c]{@{}c@{}}Top K \\ recommendation\end{tabular} &
  \multicolumn{1}{c|}{XR-Transformer} &
  \multicolumn{1}{c|}{Neighbor predictor $g$ of PINA} &
  XR-Transformer  + PINA \\ \midrule
1 &
  \multicolumn{1}{c|}{\begin{tabular}[c]{@{}c@{}}The Art of Keeping Snakes\\ (Advanced Vivarium Systems)\end{tabular}} &
  \multicolumn{1}{c|}{\textit{Guide to Owning Himalayan Cat}} &
  \begin{tabular}[c]{@{}c@{}}The Art of Keeping Snakes\\ (Advanced Vivarium Systems)\end{tabular} \\ \midrule
2 &
  \multicolumn{1}{c|}{Taming/Training Budgerigars} &
  \multicolumn{1}{c|}{\textit{\begin{tabular}[c]{@{}c@{}}Persian Cats: Everything About Purchase, \\ Care, Nutrition, Disease, and Behavior \\ (Special Chapter : Understanding Persian Cats)\end{tabular}}} &
  \textbf{Guide to Owning a Persian Cat} \\ \midrule
3 &
  \multicolumn{1}{c|}{Reptiles (1-year)} &
  \multicolumn{1}{c|}{The Art of Keeping Snakes (Advanced Vivarium Systems)} &
  \textbf{Guide to Owning Himalayan Cat} \\ \midrule
4 &
  \multicolumn{1}{c|}{\begin{tabular}[c]{@{}c@{}}A Gardener's Handbook of Plant\\ Names: Their Meanings and Origins\end{tabular}} &
  \multicolumn{1}{c|}{Chameleons (Barron's Complete Pet Owner's Manuals)} &
  Reptiles (1-year) \\ \midrule
5 &
  \multicolumn{1}{c|}{\begin{tabular}[c]{@{}c@{}}Little Dogs: Training Your\\ Pint-Sized Companion\end{tabular}} &
  \multicolumn{1}{c|}{Boas (Barron's Complete Pet Owner's Manuals)} &
  \textbf{The Complete Book of Cat Breeding} \\ \bottomrule
\end{tabular}
\end{table*}

\begin{table}[h]
\caption{Predicted results of XR-Transformer + PINA-naive on LF-Amazon-131K dataset. The setting is the same as the one used in Table~\ref{tab:exp_analysis}. None of the predicted labels match the ground truth.}
\label{tab:PINA-naive}
\small
\vspace{0.1 in}
\begin{tabular}{@{}cc@{}}
\toprule
Instance title: & \begin{tabular}[c]{@{}c@{}}Himalayan Cats \\ (Barron's Complete Pet Owner's Manuals)\end{tabular} \\ \midrule
\begin{tabular}[c]{@{}c@{}}Top K \\ recommendation\end{tabular} & XR-Transformer  + PINA-naive                                                                       \\ \midrule
1                                                               & Meditation and Its Practice                                                                        \\ \midrule
2                                                               & Science of Breath                                                                                  \\ \midrule
3                                                               & \begin{tabular}[c]{@{}c@{}}Himalayan Crystal Salt\\ Coarse Granulated - 1000 g - Salt\end{tabular} \\ \midrule
4                                                               & \begin{tabular}[c]{@{}c@{}}Original Himalayan\\ Crystal Salt Inhaler\end{tabular}                  \\ \midrule
5                                                               & \begin{tabular}[c]{@{}c@{}}Himalayan Crystal Salt\\ Fine Granulated - 1000 g - Salt\end{tabular}   \\ \bottomrule
\end{tabular}
\end{table}

% \Eli{To Jiong: Can we add some more examples here as we promise to the reviewer?}

\textbf{Qualitative analysis. }We further examine the predicted results of XR-Transformers and XR-Transformers combined with PINA (see Table~\ref{tab:exp_analysis}). The considered instance is a book about caring for Himalayan cats and the ground truth labels are books about caring for cats. Labels returned by the XR-Transformer are all related to taking care of animals or plants. However, none of them relates to cats, which is the key concept of the ground truth labels. On the other hand, XR-Transformer + PINA successfully returns all three ground truth labels within its top $5$ predictions. For further verification, we examine the predicted neighbors that are used in PINA augmentation. The top $2$ neighbors are related to keeping not only Himalayan cats but also Persian Cats, which illustrates how PINA augmentation helps the XR-Transformer to provide better predictions. Note that the PINA neighborhood predictor $g$ is trained only on plain training data, while this example was chosen from the test data. Hence, the PINA neighborhood predictor $g$ saw neither the instance title nor its associated ground truth labels. Nevertheless, it did see the label text such as ``Guide to Owning Himalayan Cat'', which is critical for training a better text vectorizer as we mentioned in the previous section. As a comparison, none of the predicted labels from XR-Transformer + PINA-naive matches the ground truth ones (Table~\ref{tab:PINA-naive}). This further demonstrates the importance of letting the text vectorizer observe the label text.

\section{Additional Experiment results}
We also examine the trained neighbor predictor $g$ on LF-Amazon131K. We find that 26\% of the predicted result from $g$ are labels (74\% are instances), and the corresponding P@1 (filtering out instances) is 33.72\%. Recall that the vanilla XR-Transformer can achieve P@1 = 45.61\%. This result shows that the gain of XR-Transformer+PINA is not merely due to being able to predict the correct labels, but the predicted instances also play an important role.

\section{Relation between PINA and the other methods}

\textbf{Comparison with~\citealt{barkan2020cold}. }In the setting of~\citealt{barkan2020cold}, the authors assume that there is a given hierarchical label tree (HLT). That is, they assume the label (tags in~\citealt{barkan2020cold}) naturally has a viable hierarchical structure. This is similar to the XR-Transformer, instead, the HLT in the XR-Transformer is built based on hierarchical k-means instead of directly given in the dataset. The key idea in~\citealt{barkan2020cold} is to “aggregate” the parent tag embeddings for all nodes in the HLT (for both items (instances) and tags) and learn them simultaneously. Indeed, \citealt{barkan2020cold} shares the similar idea of “aggregating” embeddings from others as our PINA approach. However, we would like to emphasize that the main difference between~\citealt{barkan2020cold} and our work is that~\citealt{barkan2020cold} does not aggregate predicted neighbor embeddings. Indeed, from the example given in~\citealt{barkan2020cold} (Figure 1), we can see that the “cold item” is already provided the tag information in the data. That is, we already know the parent tags of “QT8” in Figure 1-(a) and “Radio”, “Diva” in Figure 1-(b) in the case of~\citealt{barkan2020cold}. Thus, the “cold item” in~\citealt{barkan2020cold} can “aggregate” the parent tags representations. However, for a “test item” that does not have tags information (which is what we need to predict), the method in~\citealt{barkan2020cold} can only predict the relevant tags for it based on its own embeddings. No neighborhood aggregation can be applied as the relevant tags are not included in the dataset (like test labels).

\textbf{Comparison to~\citealt{wang2018billion}. }Note that in [1], the side information is naturally provided in the form of a per-item fashion. For instance, see Section 2.4 in~\citealt{wang2018billion}, where $W_v^0$ is the original item embedding of item $v$ and $W_v^i,i\geq 1$ are the side information embedding of item $v$. Thus, it is easier to infuse this side information for each item, where the authors of~\citealt{wang2018billion} take a weighted average (or concatenation) of them (see equations (6), (7), and Figure 3 in~\citealt{wang2018billion}). In contrast, both types of side information considered in our paper, label text and instance correlation signal, are not in a per-item fashion. It is impossible to directly add label text embedding to each instance without the help of our PINA formulation. Similarly, the instance correlation signal is not in the form of a per-instance (per-item) fashion, as it is of the form (bi)adjacency matrix (i.e. the instance-to-instance graph mentioned in Section 4.2). Also, note that it is prohibitive to directly concatenate the rows of the (bi)adjacency matrix to the instance features, as we potentially do not have the corresponding co-purchase information for the new instances. Thus, learning the neighbor predictor $g$ as in our PINA is crucial. Finally, while it might be possible to leverage the DeepWalk type of self-supervised loss for accounting the instance correlation signal, we would like to emphasize that it does not work for label text information. In contrast, our PINA method provides a unified way to leverage both types of side information (label text and instance correlation signal) under the XMC framework. That is, the pretraining objective is still an XMC problem so we can use the same XMC model for pretraining. This is the unique advantage of PINA in the XMC problem compared to not only~\citealt{wang2018billion} but also prior works such as SiameseXML~\cite{dahiya2021siamesexml}.

\textbf{Comparison to~\citealt{liu2021noninvasive}. }Similar to our discussion above, we note that the side information considered in~\citealt{liu2021noninvasive} is also in a per-user or per-item fashion. We would like to emphasize that the problem considered in~\citealt{liu2021noninvasive} is predicting the next item’s ID that a user $u$ would buy based on the historical action of the user 
. Roughly speaking, the problem in~\citealt{liu2021noninvasive} wants to predict a user-to-item relation and thus the user in~\citealt{liu2021noninvasive} has a similar role of instance and the item in~\citealt{liu2021noninvasive} has a similar role of the label as in the XMC problem formulation. Thus, the approach in~\citealt{liu2021noninvasive} cannot take the instance correlation signal into account (i.e., user-to-user graph in the scenario of~\citealt{liu2021noninvasive}). While both~\citealt{liu2021noninvasive} and our work consider the per-label side information, we would like to emphasize that the base model considered in~\citealt{liu2021noninvasive} and the XMC models are different. As we mentioned in section 2.3, most traditional XMC architectures treat labels as featureless identifiers. Thus, it is impossible to concatenate/append label text to labels as in~\citealt{liu2021noninvasive} for XMC models such as XR-Transformer.

\section{Pseudo Codes}\label{apx:pseudo-code}
\begin{algorithm}
  \caption{PINA augmentation}
  \label{alg:PINA_augmentation}
  \begin{algorithmic}[1]
    \STATE \textbf{Input:} neighbor predictor $g$, pretrained text vectorizer $\Phi_{\text{pre}}$, number of aggregated neighbors $K$, statistical text vectorizer $\Phi_{\text{stat}}$, instance text $\{x_i\}_{i\in [N+N_{test}]}$, label text $\{z_l\}_{l\in [L]}$.
    \FOR{$i \in [N+N_{test}]$}
      \STATE Construct ego feature $\mathbf{V}_i^{(0)} \leftarrow \Phi_{\text{pre}}(x_i)$
      \STATE Get predicted top-$K$ neighborhood prediction vector: \\ $\mathbf{P}_{i} \leftarrow top_K(g(x_i,\Phi_{\text{stat}})) \in [0,1]^{1\times (L+N)}$.
      \STATE Construct aggregated neighborhood feature: \\ $\mathbf{V}_i^{(1)} \leftarrow \sum_{j: \mathbf{P}_{ij}>0} \mathbf{P}_{ij}\Phi_{\text{pre}}(h_j)$, where $h_j = z_j$ for $j\in [L]$ and $h_j=x_{j-L}$ for $j>L$.
      \STATE Normalization: $\mathbf{V}_i^{(0)} \leftarrow \frac{\mathbf{V}_i^{(0)}}{\|\mathbf{V}_i^{(0)}\|}$, $\mathbf{V}_i^{(1)} \leftarrow \frac{\mathbf{V}_i^{(1)}}{\|\mathbf{V}_i^{(1)}\|}$.
      \STATE Concatenate ego and neighborhood feature: \\ $\mathbf{V}_i \leftarrow \left[ \mathbf{V}_i^{(0)},\mathbf{V}_i^{(1)} \right]$.
      \STATE Normalization: $\mathbf{V}_i \leftarrow \frac{\mathbf{V}_i}{\|\mathbf{V}_i\|}$.
    \ENDFOR
    \\
    \textbf{Return} PINA feature: $\mathbf{V}$.
  \end{algorithmic}
\end{algorithm}

\begin{algorithm}
  \caption{Pretraining XMC: obtaining $g$ and $\Phi_{\text{pre}}$}
  \label{alg:pretrain_xmc}
  \begin{algorithmic}[1]
    \STATE \textbf{Input:} statistical text vectorizer $\Phi_{\text{stat}}$, train instance text $\{x_i\}_{i\in [N]}$, label text $\{z_l\}_{l\in [L]}$, train multi-label $\{\mathbf{y}_i\}_{i\in [N]}$.
    \STATE Convert train multi-label to corresponding biadjacency matrix $\mathbf{B}\in\{0,1\}^{N\times L}$: $\mathbf{B}_i \leftarrow \mathbf{y}_i$.
    \STATE Construct pretraining biadjacency matrix $\mathbf{B}_{pre} = \left[ \mathbf{B}, \mathbf{I}; \mathbf{I}, \mathbf{B}^T\right].$
    \STATE Organize pretraining instances:\\ $x_{i}^{(\text{pre})} \leftarrow x_{i}\;\forall i\in[N]$, $x_{N+l}^{(\text{pre})}\leftarrow z_{l}\;\forall l\in[L]$.
    \STATE Organize pretraining labels:\\
    $\mathbf{y}_i^{(\text{pre})} \leftarrow \mathbf{B}_{pre,i}\;\forall i\in [N+L]$.
    \STATE Train XR-Transformer with \\$\{x_{i}^{(\text{pre})}\}_{i=1}^{N+L}$, $\{\Phi_{\text{stat}}(x_{i}^{(\text{pre})})\}_{i=1}^{N+L}$ and $\{\mathbf{y}_i^{(\text{pre})}\}_{i=1}^{N+L}$,\\
    resulting model $g$ and its corresponding text vectorizer $\Phi_{\text{dnn}}$.
    \STATE Concatenate dense and sparse text vectorizer: \\
    $\Phi_{\text{pre}}(\cdot) \leftarrow \left[\Phi_{\text{dnn}}(\cdot); \Phi_{\text{stat}}(\cdot)\right]$.
    \\
    \textbf{Return} Neighbor predictor $g$, pretrained text vectorizer $\Phi_{\text{pre}}$.
  \end{algorithmic}
\end{algorithm}

%%%%%%%%%%%%%%%%%%%%%%%%%%%%%%%%%%%%%%%%%%%%%%%%%%%%%%%%%%%%%%%%%%%%%%%%%%%%%%%
%%%%%%%%%%%%%%%%%%%%%%%%%%%%%%%%%%%%%%%%%%%%%%%%%%%%%%%%%%%%%%%%%%%%%%%%%%%%%%%

\end{document}